# Arabic Tweet Act: A Weighted Ensemble Pre-Trained Transformer Model for Classifying Arabic Speech Acts on Twitter

Khadejaa Alshehri, Areej Alhothali and Nahed Alowidi

**ABSTRACT**

Speech acts are a speaker's actions when performing an utterance within a conversation, such as asking, recommending, greeting, or thanking someone, expressing a thought, or making a suggestion. Understanding speech acts helps interpret the intended meaning and actions behind a speaker's or writer's words. This paper proposes a Twitter dialectal Arabic speech act classification approach based on a transformer deep learning neural network. Twitter and social media, are becoming more and more integrated into daily life. As a result, they have evolved into a vital source of information that represents the views and attitudes of their users. We proposed a BERT based weighted ensemble learning approach to integrate the advantages of various BERT models in dialectal Arabic speech acts classification. We compared the proposed model against several variants of Arabic BERT models and sequence-based models. We developed a dialectal Arabic tweet act dataset by annotating a subset of a large existing Arabic sentiment analysis dataset (ASAD) based on six speech act categories. We also evaluated the models on a previously developed Arabic Tweet Act dataset (ArSAS). To overcome the class imbalance issue commonly observed in speech act problems, a transformer-based data augmentation model was implemented to generate an equal proportion of speech act categories. The results show that the best BERT model is araBERTv2-Twitter models with a macro-averaged F1 score and an accuracy of 0.73 and 0.84, respectively. The performance improved using a BERT-based ensemble method with a 0.74 and 0.85 averaged F1 score and accuracy on our dataset, respectively.

## 1. Introduction

The essence of communication is understanding what the speaker wants to pass to the listener, the writer, and the reader. Speech act theory gives us a better understanding of speakers' intentions because it conveys an intended language's function. Speech act functions include requests, apologies, suggestions, offers, etc. The theory of speech acts was first put forth by J.L. Austin in his book How to Do Things with Words, where he emphasized that when people speak, they are also performing specific acts [1]. Knowledge of the speech acts in a text can help one to analyze the text and better understand the writer's state of mind. Analyzing speech acts in a text and improving its performance can help improve several types of applications, such as human-computer dialog, company mail or messages, and rumor detection.

CONTACT Khadejaa Alshehri. Email: kalshehri0142@stu.kau.edu.sa
Email: aalhothali@kau.edu.sa (Areej Alhothali)
Email: nalowidi@kau.edu.sa (Nahed Alowidi)

As an example, Twitter is one of the most popular social networks, and its use has increased rapidly. It is considered a communicative act due to its social networking nature. Twitter provides a channel for the public to share feelings, express their idea, request things, and ask questions. In terms of “tweet acts,” it is helpful to understand the intentions behind users’ posts, to analyze Twitter content, and to understand how users interact on social media and for which purposes they use Twitter. For example, if there are many tweets about a controversial topic, we can predict there is a misunderstanding about this topic, while if we find it tends to be expressed, we can predict there is resentment or sympathy tends to the topic. In this way, the main objective of classifying the speech act of a tweet goes beyond the literal meaning of the text and considers how the context and intention contribute to the meaning of the tweet.

Understanding what people tweet about is useful for several reasons. For example, news topics are generally meant to include facts, statements, and assertions. Therefore, if there are any deviations from our expectations – for example, if the post unexpectedly contains a product suggestion or threats to us or others – we can assume that it is spam. In addition, we can learn what individuals think about new goods and services and what worries or unnerves them. These things and more can be understood from their speeches and tweets.

Some studies have targeted classifying speech acts that appear in synchronous conver- sations as phone calls and meeting discussions, while others have focused on asynchronous conversations as tweets, emails, and forms. There have been numerous efforts to categorize speech acts in many languages, such as English [2], French [3], and Persian [4]. At the same time, little attention has been directed toward the Arabic language. Arabic is one of the Semitic languages spoken by more than 330 million people as a native language in an extended area that includes more than 20 countries [5]. The Arabic language can be characterized in three categories: classical Arabic, Modern Standard Arabic (MSA), and Arabic dialects. According to [6], classical Arabic is the language of the Holy Quran and Al-Hadith. Modern Standard Arabic (MSA) is the language of media and education in the Arab world. Finally, the Arabic dialects are the spoken languages used in daily informal communication.

The semantics of classical Arabic include statements and construction. The statement is a sentence or phrase that can be a fact (an assertion declaration that is either true or false). A construction, on the other hand, is a sentence that is not supposed to be true or false. Constructions are split into two categories, request and non-request. The request includes questions, orders, etc., while the non-request includes exclamation, praising, and complaint [7]. MSA inherited its syntax, morphology, and phonology from classical Arabic. In contrast, Arabic dialects do not follow any rules in spelling and grammar, which represents a challenge for Natural language processing (NLP) models.

As Arabic dialects are the most common type of Arabic found on social media, this paper will focus on them. There are several proposed Arabic speech act taxonomies, such as [8] and [9], but the majority of them apply to simultaneous conversations such as phone calls and meeting discussions. The other is about MSA Arabic [10] which applies rules in spelling and grammar that are not applied in dialects.

Thus, the contribution of this research is fivefold: 1) we developed a new Arabic tweet act dataset, from the ASAD dataset [11], that comprises 22, 352 tweets annotated using a domain-specific taxonomy of six speech act categories that are commonly seen on Twitter [2], 2) we implemented different variants of Arabic language Bidirectional Encoder Representations from Transformers (BERT) to classify speech acts in Arabic tweets and evaluated on the newly collected dataset, and another dataset that was previously developed in the field [12], 3) we implemented ensemble BERT model to improve and take advantage of all pre-trained BERT models in the classification of speech act, 4) we overcome the data imbalance issue using data augmentation approach based on BERT model , and 5) We provide a model explainability to identify factors that includes the model prediction and get insight into the performance of the model. In this research, we follow the taxonomy proposed by [2], which is a derivative of Searle's taxonomy [13] with a few changes to make it applicable to Twitter. The taxonomy includes six classes of speech acts: Assertion, Expression, Request, Question, Recommendation, and Miscellaneous.

The remainder of this paper is organized as follows. Section 2 outlines previous related works in speech act classification. Section 3 discusses the methodology proposed in this study. Section 4 presents performance metrics that were used to evaluate the models. Section 5 expounds on the performance results of the proposed models. Finally, the Model Performance Interpretation and conclusions are presented in Sections 6 and 7, respectively.

## 2. Related Work

Speech act classification, or analyzing a user's speech act inside a conversation, is a recent active study subject in natural language understanding. Speech act classification has been used for a variety of purposes, including detecting rumors [14], automatic tweet topic summarization [15], crisis response from social media [3], and political campaign message classification [16]. Most researchers classify the dialogue act, which is considered a synchronized conversation [17–19]. Others try to classify tweet act [2,10,20–26] , or discussion forms [27], which are considered asynchronous conversations.

Zhangetal. [21]builtanSupportvectormachine(SVM)classifierwithalinearkernel to classify a manually annotated dataset (8613 tweets) into one of five proposed speech act categories. They proposed a variety of features based on words and characters that have been handcrafted. Their method obtained a weighted average F1 score of nearly 0.70. Meanwhile, in [2], Vosoughi et al. constructed a classifier that categorizes tweets into one of six speech acts. They developed a logistic regression classifier with over 3, 000 binary semantic and syntactic features. Their classifier provided a weighted average F1 score of 0.70. Their classifier was used to detect rumor tweets in Vosoughi et al. [14] dataset.

Recently, Saha et al. [20] presented the first deep-learning classifier. They developed a Convolutional Neural Network (CNN) model with different activation functions, such as softmax and linear support vector, at the last layer. They also utilizes several handcrafted features, such as emotions, opening words, and N-gram. To overcome the out-of-vocabulary problem, they used pertained GloVe embedding for input words and characters. Using a 7,000-tweet dataset that was manually annotated with seven proposed speech acts, they achieved 73.75% accuracy and a 71% F1 score for the best model, which is the CNN-SVM model.

Using the dataset published by Saha et al. [20], the authors in [20] proposed a BERT- Extended-based speech act classifier. They compared their work with several baseline models, such as LSTM, BiLSTM, and CNN- LSTM. They obtained an accuracy and F1 score of 75.97% and 74%, respectively. Later, in [23], they developed a BERT-Caps model based on BERT. Unlike previous work, they included syntactic and semantic features as binary features. For comparison, they built several baseline models. Their model achieved 77.52% accuracy and 0.77 weighted average F1 score.

As mentioned previously, little attention has been directed to the Arabic language. As far as we know, only four studies have focused on the Arabic speech act classification task. Sherkawi et al. [24] proposed two main classification approaches: the rule-based expert system and machine-learning methods. The rule- based expert system was constructed using a bootstrapping method based on the notion that Arabic language grammar is naturally rule-based. For the second approach, they investigated different machine learning methods, such as Decision Trees, Naive Bayes, Neural Networks, and SVM. The rule-based system relies on Arabic grammar, whereas machine learning relies on several types of features. They proposed a taxonomy of 16 speech acts derived from Arabic rhetoric science. For evaluation, they built a dataset of 1, 500 sentences written in MSA. They obtained an accuracy rate of 98.92% for the rule-based expert system and up to 97.09% for the machine- learning approaches. However, their taxonomy is suitable for MSA only, and the dataset is too small.

The second study [25] used the Arabic Speech-Act and Sentiment Corpus of Tweets (ArSAS) [12], which is the first corpus that has speech act labels. It follows the taxonomy proposed in [2]. They developed two different approaches to classify the Arabic tweet act. The first approach was an SVM classifier with a set of syntactic and semantic features. The second approach applied several deep-learning methods, such as CNN, LSTM, BiLSTM, and several combinations between them. They represented the word in each tweet using the skip-gram word2vec embedding. For the SVM with all proposed features, they obtained an accuracy of up to 86.5% and a weighted average F1 score of 86.0%. However, for deep- learning methods, they found that the BiLSTM has the best predictive performance with 87.5% accuracy and 86.0% weighted average F1 score.

The third study [10] was based on the previous study, but they enhanced the dataset with extra annotated features, such as emoticons, links, sentiment labels, and tweet length. Using these features with the SVM classifier, the researcher improved the weighted average F1 score to 89.0% compared with the SVM classifier with features proposed in [25]. In the last study [26], they evaluated different pre-trained BERT models using two public datasets, namely, ArSAS and levInt. The best results were archived using the AraBERTv0.2-base model with 89% and 94% accuracy, respectively.

To the best of our knowledge, there is only one public dataset for Arabic Speech Act on Twitter, which is ArSAS [12]. The ArSAS collection uses 20 topics, which tends to make it contain a lot of controversial topics, hashtags, and links. Also, it tends to be a high-imbalance dataset. ArSAS limitations encourage us to annotate new datasets that were collected in a more general way without any specific topic. in In our work, we introduce a new dataset of Arabic speech acts. We evaluated it using the two proposed models and compared the results with ArSAS2. The following section presents our annotation process with extensive descriptions.

## 3. Methodology

This section describes the dataset and its preprocessing steps, as well as the classification models that are used.

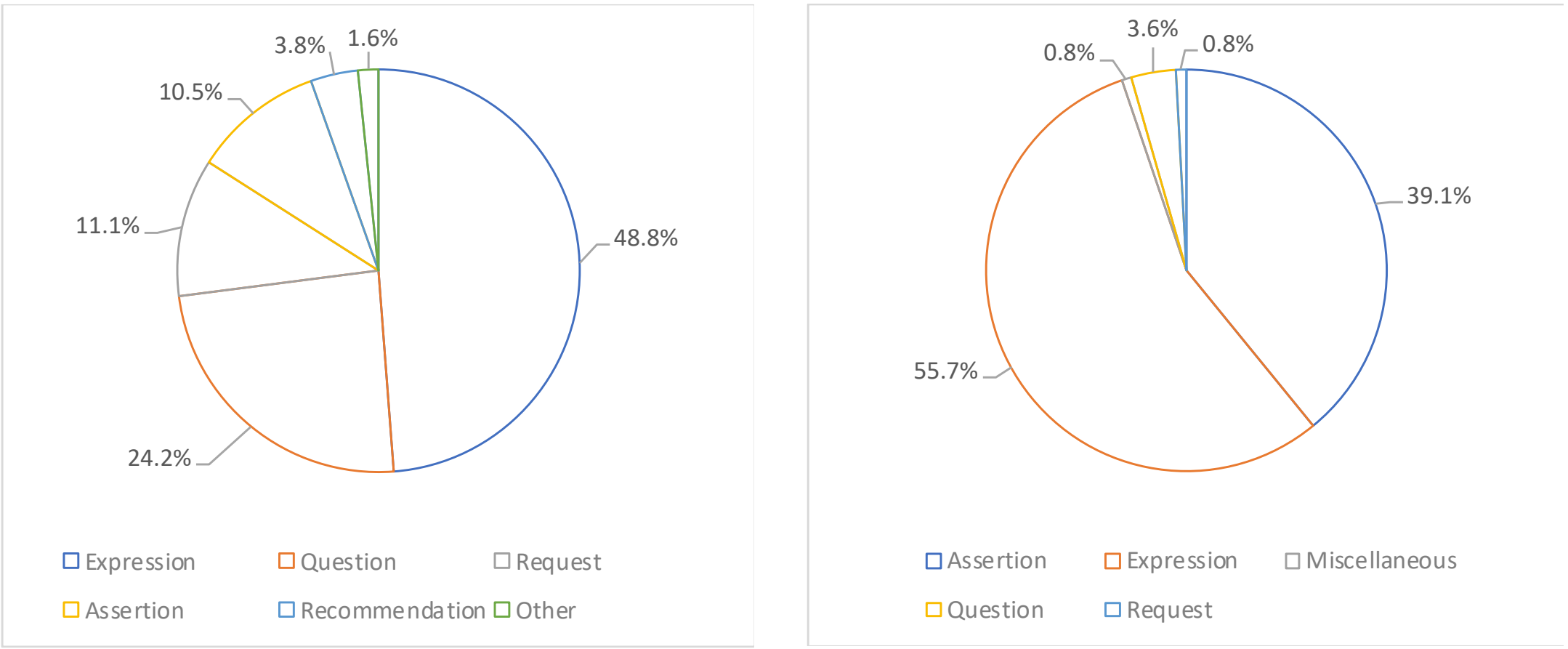

*(a) Distribution of speech act classes in the ASAD dataset*

*(b) Distribution of speech act classes in the ArSAS dataset*

*Figure 1 Distribution of speech act classes in ASAD and ArSAS datasets*

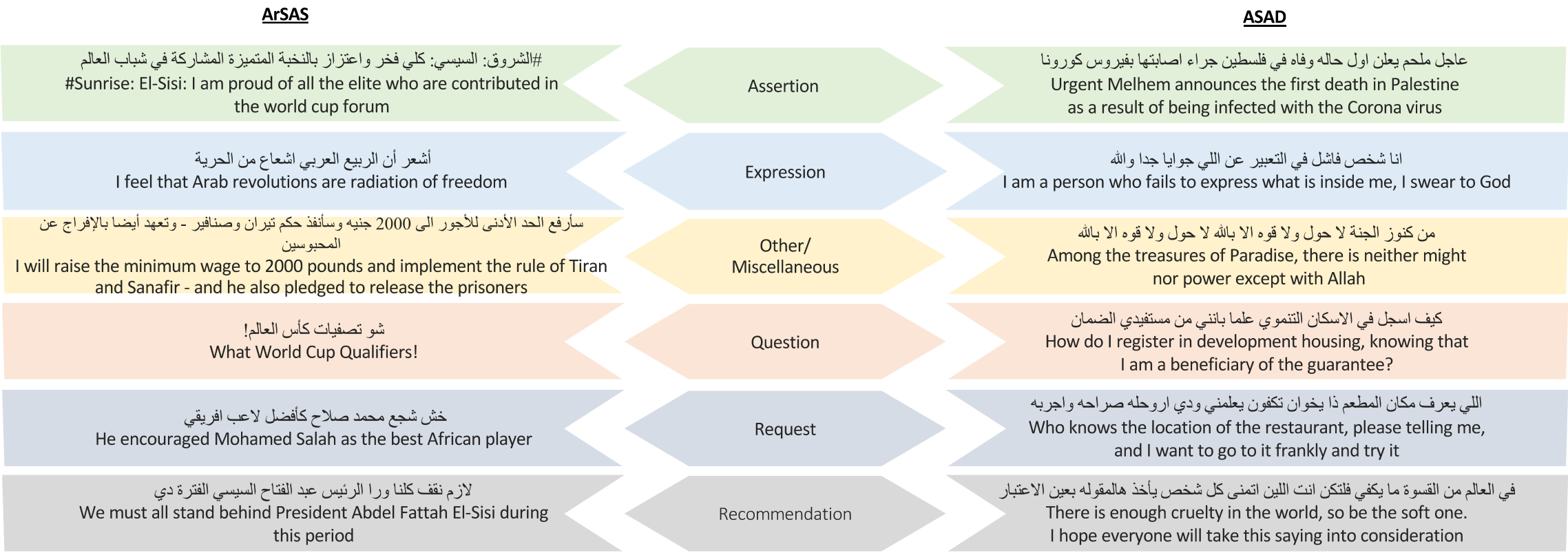

*Figure 2 Sample tweets of different speech act categories in the ArSAS and ASAD datasets*

## *3.1. Dataset*

We used a part of the public ASAD dataset [11] for sentiment analysis tasks. The ASAD is available for researchers to accelerate Arabic research in NLP [11]. The ASAD was collected using Twitter API between May 2012 and April 2020. In the selection process, they used several filtration operations to ensure that the Arabic texts were sufficient for the annotation task. First, they selected a random set of tweets that has the label "ar" from the pool and added it to the corpus. Then, they removed the tweets that contained less than seven words. Finally, they obtained a corpus of 100, 000 tweets. Since the corpus was published with only sentiment analysis, we have annotated a total of 25, 000 of ASAD using a domain-specific taxonomy of six speech act categories that are commonly seen on Twitter [2] similar to the speech act labels of the ArSAS dataset.

Google Forms were used as an annotation platform for the ASAD dataset with six speech act labels, given that there is no annotation platform that supports Arabic annotation. To make it easier for the annotators, the ASAD tweets were divided into 100 tweets per form. Each form was answered by three annotators. All annotators have the Arabic language as their mother tongue, and their ages are between 16–60 years. The annotators were given a clear set of guidelines, as well as examples of tweets from each speech act class, to help them understand the nature of these classes. However, some tweets were annotated with more than one label. To solve this problem, these tweets were labeled based on majority voting. Tweets that had different annotation by the judges were excluded. Finally, we came up with a corpus of 22, 352 tweets. Figures 1a and 1b present the distribution of speech act classes in the ASAD and ArSAS datasets, respectively, and Figure 2 shows examples of tweets with different speech act categories in the ASAD dataset.

### *3.2. Data Splitting*

The total number of tweets in the newly annotated dataset, ASAD, is 22, 352. We followed the "80,20" principle to divide our data into training and testing sets, where 20% of the samples from each class must be present in the test set.

### *3.3. Data Preprocessing*

Several preprocessing and normalization methods were performed on the dataset to improve overall predictive results. The following normalization steps were applied:

- Change all URLs to the "URL" tag.
- Change the hashtag symbol to "hash" tag.
- Change the mention symbol to "user" tag.
- Change the question and exclamation symbols to "question" and "exclamation" tags, respectively.

- Remove all diacritics, punctuations, non-Arabic characters, emojis, new lines, and any extra spaces.
- normalise all letters that appear in different forms into a single form. As an example, (إأآ) to (ا).
- Remove longation; any character that appeared more than twice was removed.

Finally, after applying all the previous steps, we removed any tweet that had less than three words.

### *3.4. Data Augmentation*

Data augmentation is a technique that helps to regularize the model and address various challenges. One of these challenges is the imbalance of the dataset, which can be solved by data augmentation. The idea is to create new samples from the existing ones using paraphrasing approch or . This way, the data is enriched and more diverse. We can see from Figures 1a and 1b that both ASAD and ArSAS are imbalanced datasets. To balance them, we increased the size of all classes—except the largest one—to match the largest one. We first divided the dataset into training and testing sets, and then applied data augmentation to each set separately using the NLPAUG library [28]. We used BertAug, which uses the BERT model, to insert new words based on contextual word embeddings. Figures 3a and 3b show the distribution of Speech Act classes in the augmented ASAD and ArSAS datasets.

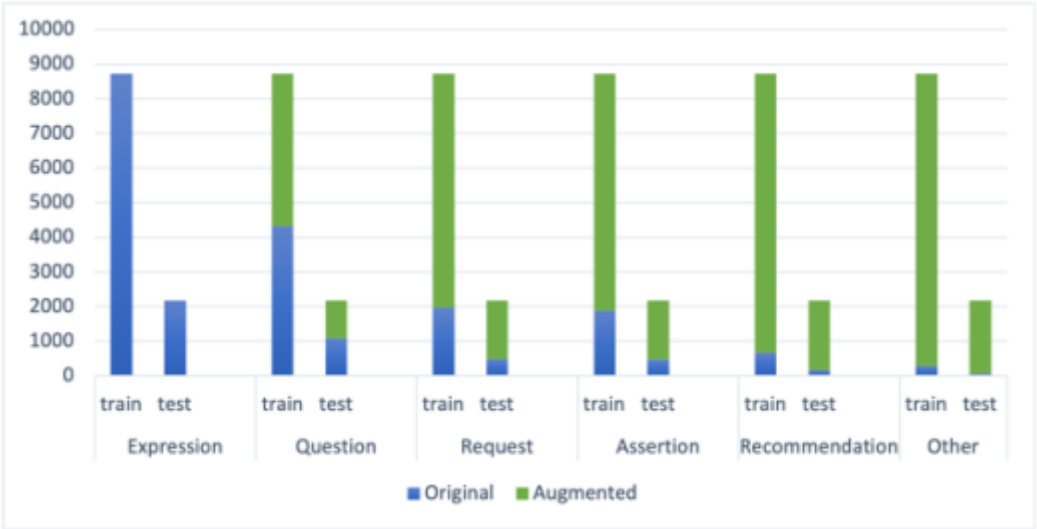

**(a)** Data augmentation distribution of Speech Act classes in the ASAD dataset.

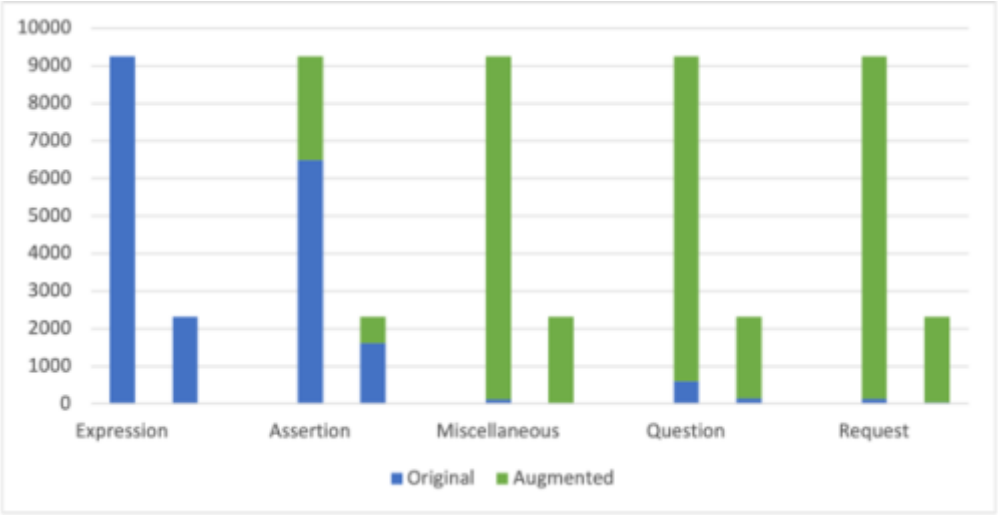

**(b)** Data augmentation distribution of Speech Act classes in the ArSAS dataset.

*Figure 3 Data augmentation distribution of Speech Act classes in the ASAD and ArSAS datasets*

### *3.5. Classification Model*

In this paper, we implemented two types of deep neural network models. The first model is BiLSTM, with AraVec as word embedding. The second model is The Bidirectional Encoder Representations from Transformers (BERT) [29]. BERT is a language representation model introduced recently that has produced cutting-edge outcomes for a wide range of NLP tasks. We utilized in this study several BERT models for the Arabic language, with a fine-tuning and classification output layer.

#### 3.5.1. BiLSTM Experimental Settings

In this section, a deep-learning model that has been tested consists of two BiLSTM hidden layers, followed by a fully connected layer with a softmax output layer activation function. The word vector used in this experiment was embedded using an Arabic pre- trained word embedding that has been utilized for this work called "AraVec" [30] with an embedding size of 300. The batch size was set to 50, and the dropout ratio was set to 0.50. The cost function was the sparse categorical cross-entropy, which measures the difference between the true and predicted labels. The optimizer was ADAM, which is an adaptive learning rate method that adjusts the parameters based on the gradients. The epoch size was set to 15, which means that the model trains for 15 iterations over the entire data set. The LSTM unit size was 100, which means that the model has 100 memory cells in each LSTM layer.

3.5.2. BERT Experimental Settings

We used the Simple Transformers library to fetch the pre-trained BERT models. The library provided pre-trained BERT models by name. The different versions of the BERT models that were tested include bert-base-arabertv01, bert-base-arabertv02, and bert-base- arabertv02-Twitter provided by [31], bert-base-Arabic-camelbert-mix provided by [32], MARBERTV2 by [33], and bert-base-qarib by [34]. We fine-tune all models with the default hyper-parameter values, except for the epoch number and learning rate, which are set to 4 and 4e-05, respectively.

3.5.3. Model Ensemble

The idea behind an ensemble method is to combine multiple different models to obtain better performance than could be obtained from any of them alone. In our work, we combine pre-trained BERT models that use tweets in the training process, which are bert-base-arabertv02-Twitter [31], bert-base-Arabic-camelbert-mix [32], MARBERTV2 [33], and bert-base-qarib [34] and exclude others.

As shown in Figure 4, the data were fed to pre-trained BERT models. Then we summed up all the prediction probabilities. After that, we use the Argmax function to generate the final prediction.

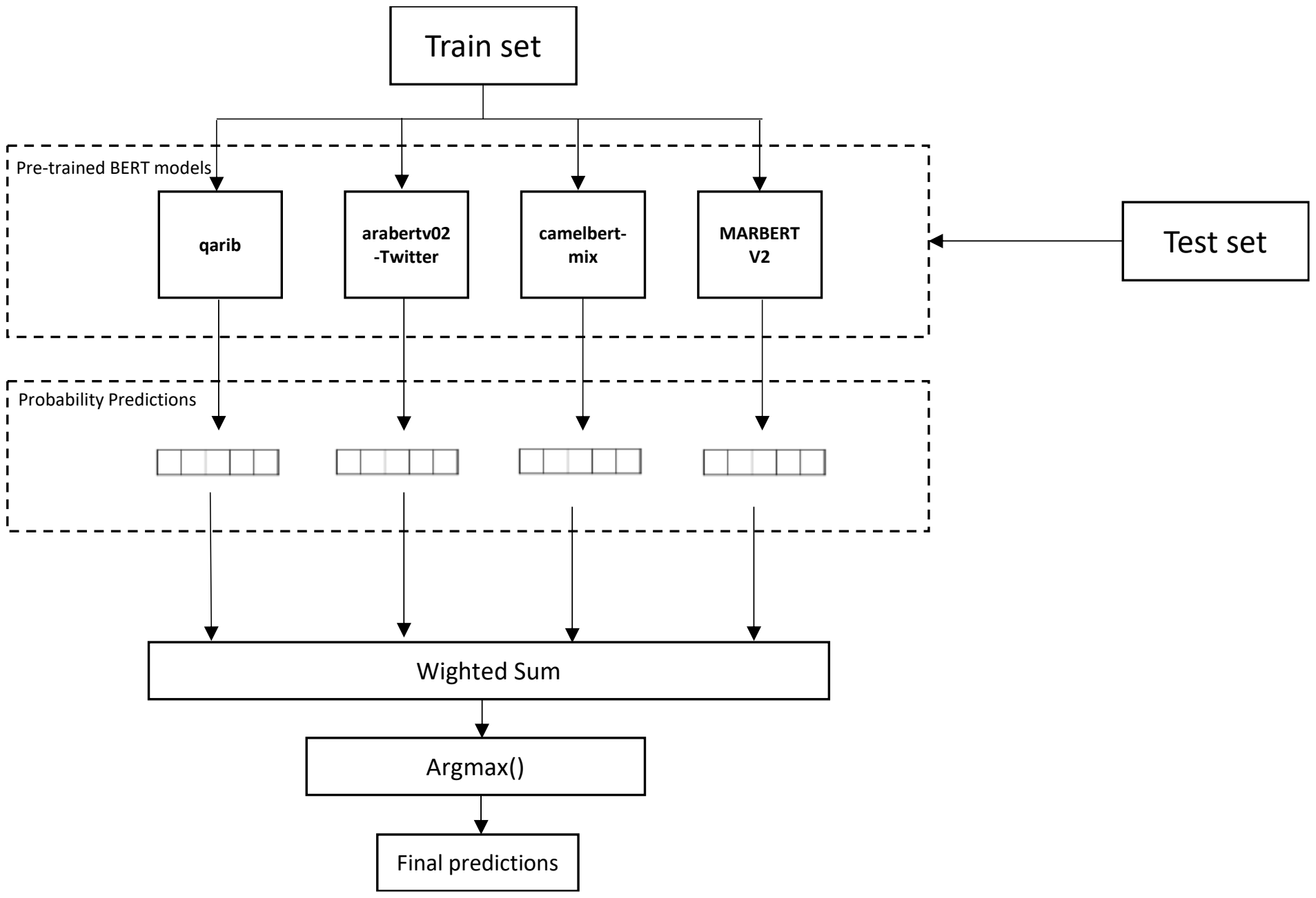

*Figure 4 Ensemble Model Architecture*

## 4. Metrics

Different performance metrics, including accuracy, precision, recall, and the F1 score (F1), are used to measure the results of the experiments with the BiLSTM and BERT models, as shown in the following equations [35].

$$Accuracy = \frac{number\ of\ correct\ predictions}{total\ number\ of\ input\ sample}$$

$$Precision = \frac{True\ Positives(TP)}{True\ Positives(TP) + \ False\ Positives(FP)}$$

$$Recall = \frac{True\ Positives(TP)}{True\ Positives(TP) + False\ Negatives(FN)}$$

$$F1 = \frac{2 * Precision * Recall}{Precision + Recall}$$

*Table 1 The overall performance of BiLSTM and BERT models on ASAD and ArSAS in terms of precision (P), recall (R) and f1-score (F1).*

| | **BiLSTM** | | | **arabertv01** | | | **arabertv02** | | | **arabertv02-Twitter** | | | **camelbert-mix** | | | **MARBERTV2** | | | **qarib** | | |
|---|---|---|---|---|---|---|---|---|---|---|---|---|---|---|---|---|---|---|---|---|---|
| **ASAD** | | | | | | | | | | | | | | | | | | | | | |
| | **P** | **R** | **F1** | **P** | **R** | **F1** | **P** | **R** | **F1** | **P** | **R** | **F1** | **P** | **R** | **F1** | **P** | **R** | **F1** | **P** | **R** | **F1** |
| **Exp** | 0.82 | 0.88 | 0.85 | 0.84 | 0.86 | 0.85 | 0.87 | 0.85 | 0.86 | 0.87 | 0.86 | **0.87** | 0.85 | 0.87 | 0.86 | 0.87 | 0.87 | **0.87** | 0.86 | 0.86 | 0.86 |
| **Que** | 0.89 | 0.89 | 0.89 | 0.90 | 0.90 | 0.90 | 0.88 | 0.92 | 0.90 | 0.90 | 0.91 | **0.91** | 0.89 | 0.91 | 0.90 | 0.90 | 0.90 | 0.90 | 0.90 | 0.90 | 0.90 |
| **Req** | 0.75 | 0.67 | 0.71 | 0.74 | 0.74 | 0.74 | 0.75 | 0.81 | **0.78** | 0.74 | 0.82 | **0.78** | 0.75 | 0.75 | 0.75 | 0.74 | 0.78 | 0.76 | 0.75 | 0.77 | 0.76 |
| **Ass** | 0.80 | 0.77 | 0.79 | 0.75 | 0.77 | 0.76 | 0.77 | 0.80 | 0.78 | 0.80 | 0.81 | **0.81** | 0.80 | 0.78 | 0.79 | 0.79 | 0.81 | 0.80 | 0.77 | 0.8 | 0.78 |
| **Rec** | 0.46 | 0.36 | 0.40 | 0.46 | 0.41 | 0.43 | 0.56 | 0.51 | 0.53 | 0.53 | 0.55 | **0.54** | 0.56 | 0.43 | 0.48 | 0.51 | 0.49 | 0.50 | 0.52 | 0.46 | 0.49 |
| **Oth** | 0.62 | 0.32 | 0.42 | 0.60 | 0.29 | 0.39 | 0.73 | 0.38 | **0.50** | 0.75 | 0.38 | **0.50** | 0.63 | 0.33 | 0.44 | 0.67 | 0.36 | 0.47 | 0.58 | 0.4 | 0.48 |
| **ArSAS** | | | | | | | | | | | | | | | | | | | | | |
| | **P** | **R** | **F1** | **P** | **R** | **F1** | **P** | **R** | **F1** | **P** | **R** | **F1** | **P** | **R** | **F1** | **P** | **R** | **F1** | **P** | **R** | **F1** |
| **Ass** | 0.86 | 0.92 | 0.89 | 0.88 | 0.90 | 0.89 | 0.90 | 0.91 | **0.90** | 0.89 | 0.90 | **0.90** | 0.89 | 0.89 | 0.89 | 0.89 | 0.90 | 0.89 | 0.89 | 0.89 | 0.89 |
| **Exp** | 0.90 | 0.89 | 0.89 | 0.89 | 0.89 | 0.89 | 0.90 | 0.90 | **0.90** | 0.90 | 0.91 | **0.90** | 0.89 | 0.91 | **0.90** | 0.89 | 0.90 | **0.90** | 0.89 | 0.90 | **0.90** |
| **Mis** | 0.25 | 0.06 | 0.10 | 0.20 | 0.09 | 0.13 | 0.18 | 0.12 | 0.15 | 0.11 | 0.06 | 0.08 | 0.23 | 0.22 | **0.23** | 0.04 | 0.03 | 0.04 | 0.10 | 0.09 | 0.10 |
| **Que** | 0.58 | 0.41 | 0.48 | 0.54 | 0.49 | 0.52 | 0.59 | 0.5 | 0.54 | 0.58 | 0.51 | 0.54 | 0.62 | 0.5 | **0.56** | 0.61 | 0.51 | 0.55 | 0.56 | 0.49 | 0.53 |
| **Req** | 0.5 | 0.37 | 0.34 | 0.32 | 0.31 | 0.32 | 0.35 | 0.4 | 0.37 | 0.38 | 0.31 | 0.34 | 0.43 | 0.29 | 0.34 | 0.38 | 0.34 | 0.36 | 0.44 | 0.34 | **0.39** |

# 5. Results

The outcomes of the experiments utilizing the BiLSTM and BERT models are presented and analyzed in this section. In the first section, the results of the experiment with the ASAD dataset are shown, followed by the comparison experiment results with the public ArSAS dataset.

## *5.1. ASAD*

Table 1 shows the outcome of applying BiLSTM with araVec word embedding to the ASAD dataset. The experiment result shows that the Recommendation class achieved the worst result in terms of F1 score by 0.40, while the Question class achieved the best by 0.89, followed by the Expression, Assertion, Request, and finally, the other class by 0.85, 0.79, 0.71, and 0.42, respectively. As presented in Table 2, the overall result is 0.68 in terms of macro avg F1, 0.82 in terms of accuracy, and 0.81 in terms of weighted average. Regarding the BERT models, we evaluated different pre-trained versions of the BERT models and fine- tuned them for Arabic speech act classification tasks using the transformers library. Table 1 displays the various BERT model versions that were tested on the ASAD dataset. All models achieved good classification results. The model bert-base-arabertv02-Twitter performed the best in our task and improved the results of all classes individually. Moreover, the overall result was improved to 0.84, 0.73, and 0.84 for accuracy, macro-average F1, and weighted-average F1, respectively, as shown in Table 2.

We used 'bert-base-arabertv02-Twitter', 'bert-base-Arabic-camelbert-mix', 'MARBERTV2', and 'bert-base-qarib' as combination BERT models in our ensemble model. As you can see in Table 2, the BERT-based ensemble model has the highest values for all three metrics, indicating that it is the best-performing model on the ASAD dataset. The BERT-based ensemble model outperforms the other models, and this shows the benefit of combining different BERT models since they use a different tweet corpus that was collected from different sources and time periods. The tweet corpus of each model is also different in size, content, and language variant. Therefore, the four models have different characteristics and capabilities when it comes to handling Arabic and dialectal text.

An augmentation technique was used to deal with the class imbalance issue by employing the best BERT model 'bert-base-arabertv02-Twitter'. The macro average F1 score for the two least frequent classes (Rec and Oth) improved considerably when we used the augmented data, as shown in Table 3a, compared to Table 1 without augmentation. That boosted the overall F1 score but reduced the other metrics. It is expected because the imbalanced dataset tends to classify the more common classes better than the less common ones.

*Table 2 The overall performance of BiLSTM And BERT models on ASAD and ArSAS in terms of accuracy (Acc), macro average f1 (M-F1), and weighted average f1 (W-F1).*

| | ASAD | | | ArSAS | | |
|---|---|---|---|---|---|---|
| | Acc | M-F1 | W-F1 | Acc | M-F1 | W-F1 |
| BiLSTM | 0.82 | 0.68 | 0.81 | 0.87 | 0.56 | 0.87 |
| arabertv01 | 0.82 | 0.68 | 0.82 | 0.87 | 0.55 | 0.87 |
| arabertv02 | 0.84 | 0.73 | 0.83 | **0.88** | 0.57 | **0.88** |
| arabertv02-Twitter | **0.84** | **0.73** | **0.84** | **0.88** | 0.55 | **0.88** |
| camelbert-mix | 0.83 | 0.7 | 0.83 | **0.88** | **0.58** | 0.87 |
| MARBERTV2 | 0.84 | 0.72 | 0.84 | 0.87 | 0.55 | 0.87 |
| qarib | 0.83 | 0.71 | 0.83 | 0.87 | 0.56 | 0.87 |
| BERT-based ensemble | **0.85** | **0.74** | **0.85** | **0.89** | **0.58** | **0.88** |

*Table 3 Best performing model (arabertv02-Twitter and camelbert for augmented ASAD and augmented ArSAS, respectively) in terms of precision(P), recall(R), f1-score(F1) and Accurcy.*

*(a) arabertv02-Twitter for augmented ASAD*

| | **P** | **R** | **F1** |
|---|---|---|---|
| **Exp** | 0.85 | 0.89 | 0.87 |
| **Que** | 0.82 | 0.91 | 0.86 |
| **Req** | 0.75 | 0.81 | 0.78 |
| **Ass** | 0.78 | 0.81 | 0.79 |
| **Rec** | 0.76 | 0.76 | 0.76 |
| **Oth** | 0.85 | 0.62 | 0.72 |
| **Acc** | | | 0.80 |
| **M-F1** | | | 0.80 |
| **W-F1** | | | 0.80 |

*(b) camelbert-mix on augmented ArSAS*

| | **P** | **R** | **F1** |
|---|---|---|---|
| **Ass** | 0.52 | 0.92 | 0.66 |
| **Exp** | 0.88 | 0.92 | 0.90 |
| **Mis** | 0.69 | 0.36 | 0.47 |
| **Que** | 0.77 | 0.83 | 0.80 |
| **Req** | 0.82 | 0.50 | 0.62 |
| **Acc** | | | 0.70 |
| **M-F1** | | | 0.69 |
| **W-F1** | | | 0.69 |

We tested our models against the publicly available ArSAS dataset to ensure that they performed well with our dataset. The next section describes the ArSAS in general, then presents the results with the BiLSTM model and various BERT model versions that were tested on the ASAD dataset.

*5.2. ArSAS*

ArSASisapublicdatasetcollectedandpresentedbyElmadanyetal. [12].ArSAS tweets are collected using the Twitter API by a set of topics. A topic is a subject discussed in one tweet or more. They used the definitions of three different types of topics proposed by [36]:

- long-standing; a topic that is discussed over a long period like أزمة الخليج ، الربيع العربي which means Arab spring , Gulf crisis, respectively.
- Entity; a topic about celebrities or organizations like ولي العهد السعودي ، الحوثيون which means KSA crown prince, Houthis, respectively.
- Event; a topic about an important thing that happened like تصفيات كأس العالم ، منتدى الشباب العربي which means World cup qualifications , World youth forum, respectively.

After collecting and cleaning the data, they annotated tweets with two labels—one for speech act (Assertion, Expression, Request, Question, Recommendation, and Miscellaneous) and the other for sentiment analysis. Figure 1b presents the distribution of speech act classes in the ArSAS dataset, and Figure 2 shows examples of tweets with different speech act categories in the ArSAS dataset. The data preprocessing steps described in Section 3.3 were used on the ArSAS dataset. Moreover, we merged the two smallest classes, which are Recommended and Miscellaneous, into one class named Miscellaneous.

Table 1 reports the result obtained when BiLSTM with araVec word embedding was applied. We observed that the assertion and expression classes were classified correctly compared with the other small classes, such as miscellaneous, requests, and questions. This could be explained as a result of the high imbalance issue in the size of the classes in the dataset. The overall result of the model is 0.56 in terms of the macro-average F1 score and 0.87 in terms of accuracy and weighted average F1, respectively, as presented in Table 2.

Table 1 displays the performance of the various BERT model versions in each of the classes. All models perform well with the two large classes, Assertion and Expression, with an F1 score between 0.89 and 0.90, compared with the other three small classes. The camel-mix model produced the best F1 scores of 0.23 and 0.56 for the Miscellaneous and Question classes, respectively, while the Request class obtained a 0.39 F1 score using the qarib model. As presented in Table 2, all models achieved good classification results. With an accuracy of 88%, the arabertv02, arabertv02-Twitter, and camelbert-mix models outperform the others. The camelbert-mix model has the best classification performance of all models, with 0.58 F1-scores calculated using the macro average. As we did with ASAD, we used 'bert-base-arabertv02-Twitter', 'bert-base-Arabic-camelbert-mix', 'MARBERTV2', and 'bert-base-qarib' as combination BERT models in our ensemble model. As we can see in Table 2, the differences between the models are not very large. However, BERT-based ensembles improve accuracy to 89%.

To overcome the limitations of the class imbalance, the augmentation technique was used. To the best of our knowledge, we are the first to test the augmentation technique with the public ArSAS dataset. As presented in Table 3b, we evaluate augmentated ArSAS using the 'camelbert-mix' model. The F1 score was improved for all three smallest classes compared without using the data augmented in Table 1. Also, the Overall F1 score was highly improved.

Table 4 presents a performance comparison with the state-of-the-art. We obtained results similar to those obtained by [26] using arabertv02. Since the ArSAS dataset does not have a public split set for training and testing, we cannot know how they split their data. In our case, we use the same split by setting the seed of 42. Using the camelbert-mix model, our average macro F1 score outperforms [26] by 5%. In addition, the average macro F1 improved to 0.69 with the augmented ArSAS dataset.

*Table 4 Performance comparison with the state of the art on ArSAS*

| | Acc | M-F1 | W-F1 |
|---|---|---|---|
| **AraBERTv0.2[26]** | 0.89 | 0.53 | 0.89 |
| **AraBERTv0.2** | 0.88 | 0.57 | 0.88 |
| **AraBERTv0.2Twitter** | 0.88 | 0.55 | 0.88 |
| **camelbert-mix** | 0.88 | **0.58** | 0.87 |
| **BERT-based ensemble** | **0.89** | 0.57 | **0.88** |
| **camelbert-mix (with augmented ArSAS)** | 0.70 | **0.69** | 0.69 |

## 6. Model Performance Interpretation

In this section, we present an analysis and interpretation of the results that have been obtained using the best BERT model on the ASAD and ArSAS datasets.

## 6.1. ASAD

By creating the confusion matrix in Table 5, we were able to analyze the classification results that the best model, “bert- base-arabertv02-Twitter,” had made. We observed that all the misclassification results with the expression class, while the highest number of misclassifications were made in tweets that belonged to the class expression that was predicted as a request (with 95 cases out of 2,182). This could be the cause of the dataset imbalance problem, which biased the model toward the majority class. Table 6 displays some misclassified examples that tend to be classified as expressions. As we can see and deduce from the tweets, the model’s predicted result may overlap with the correct speech act. This is an example of an Arabic speech act issue.

*Table 5 Confusion matrix for the best model “bert-base-arabertv02-Twitter” on the ASAD*

| | **Exp** | **Que** | **Req** | **Ass** | **Rec** | **Oth** |
|---|---|---|---|---|---|---|
| **Exp** | *1873* | 79 | **95** | 65 | 64 | 5 |
| **Que** | **61** | *985* | 26 | 8 | 1 | 0 |
| **Req** | **48** | 21 | *407* | 4 | 11 | 4 |
| **Ass** | **75** | 4 | 6 | *380* | 5 | 0 |
| **Rec** | **57** | 3 | 9 | 7 | *93* | 0 |
| **Oth** | **28** | 1 | 7 | 9 | 0 | *27* |

*Table 6 Examples of ASAD classification errors*

| **Tweet** | **Orig SA** | **pred SA** |
|---|---|---|
| بعد موت الماره في شوارع تركيا بالكورنا الان تجتاح السجون هناك رابط | **Ass** | **Exp** |
| مستخدم لااحد يجيب علي الخاص ولا احد يجيب هنا اين نذهب | **Qus** | **Exp** |
| مستخدم تقدمت بطلب تصريح قبل حمسه ايام ولا جاني اي رد بالقبول او الرفض تضرر حلالنا من التاخير رقم التصريح | **Req** | **Exp** |
| هي دي الصيغه الحقيقيه للخبر الخبر لازم يتقال كده | **Rec** | **Exp** |
| اشمعنا انا محدش قالي كده ولا انا صباره رابط | **Other** | **Exp** |

To understand more about errors, we applied Local Interpretable Model-Agnostic Explanations (LIME) to provide explanations for which words have an impact on the predicted result [37]. We presented different scenarios for sentence predictions for each label. We concentrated on four examples from the testing dataset. A screenshot of LIME explanation is shown in Figures 5a to 5d. The prediction probability for each class is shown in the left panel. The right panel displays the absolute values of the seven largest interpretable coefficients as a bar chart. The bottom panel presents the tweet to explain where the words deemed important for the prediction are highlighted. Figures 5a and 5b show the correct prediction for two tweets from the Question and Request classes. Some common words belonging to these classes are highlighted. On the other hand, Figures 5c and 5d show the incorrect predictions for the Recommendation and Expression classes. This indicate the challenge of predicting a single speech act when the sentence contains multiple acts of speech. As illustrated in Figure 5c, the sentence begins with a question, then a recommendation, and finally, predicted as an expression. Figure 5d presents how the model oscillates between the correct and wrong classes and then nominates one of them based on the highlighted words.

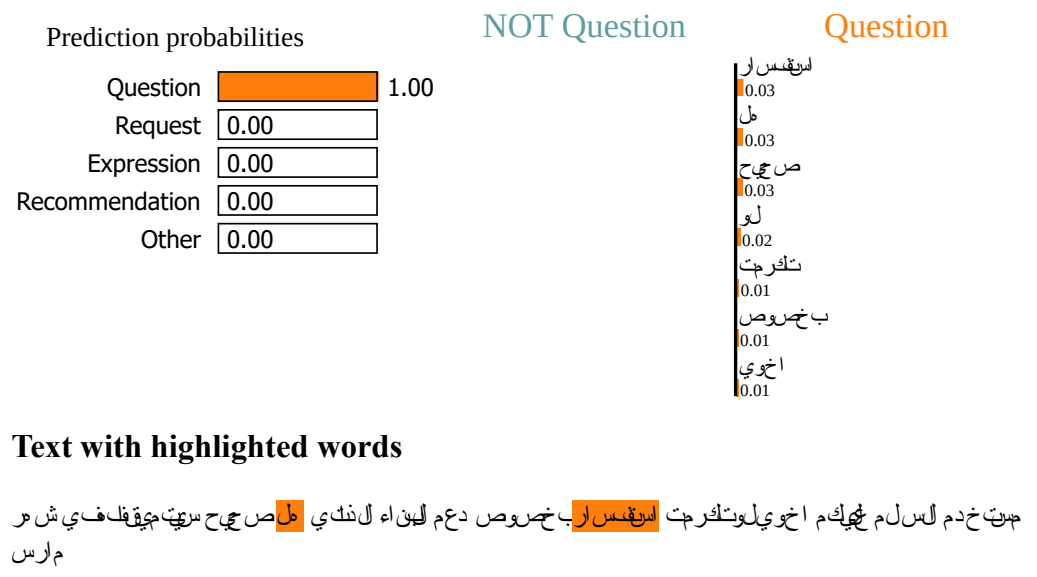

**(a)** The actual label "Question" and predicted as "Question"

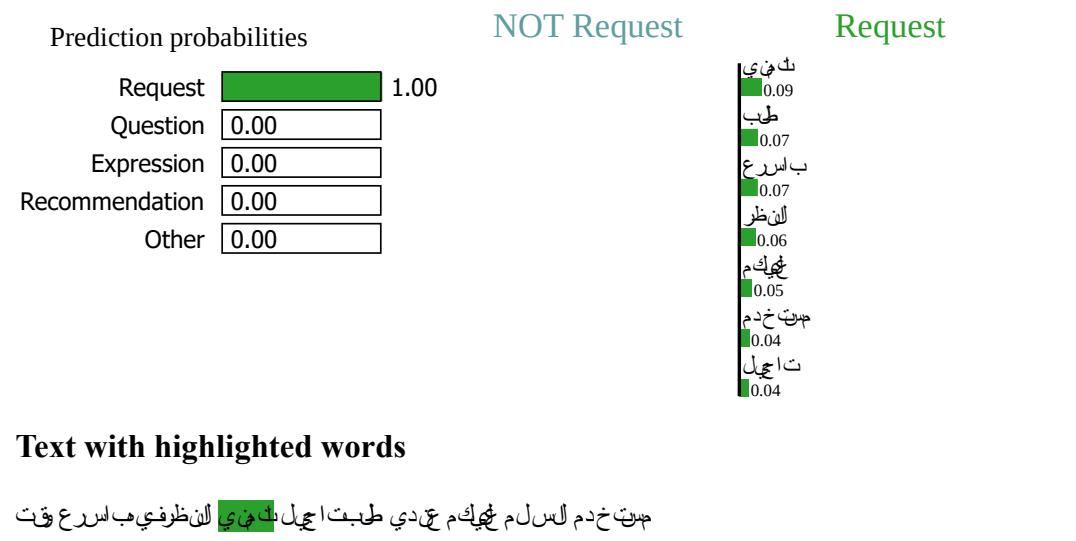

**(b)** The actual label "Request" and predicted as "Request"

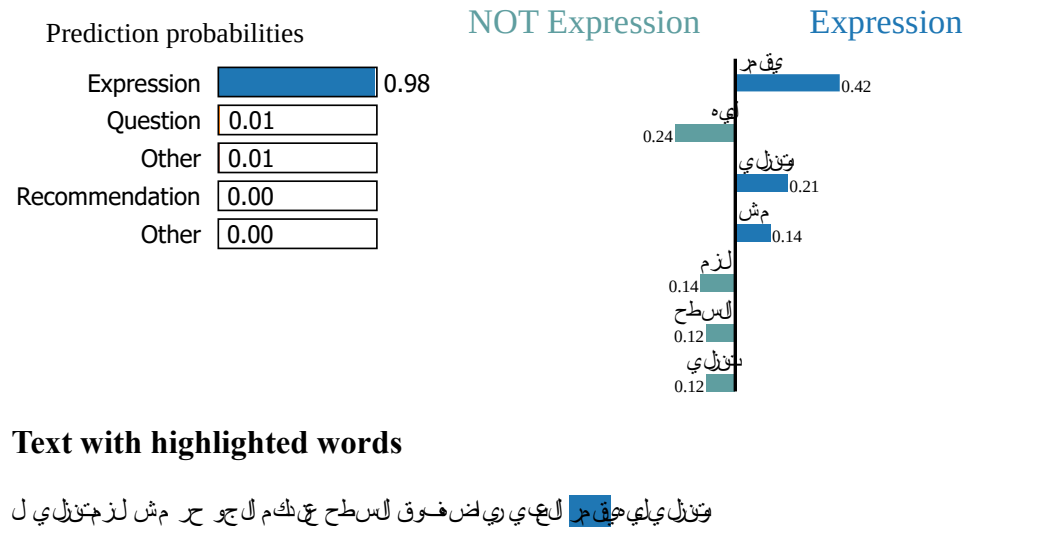

**(c)** The actual label "Recommendation" and predicted as "Expression"

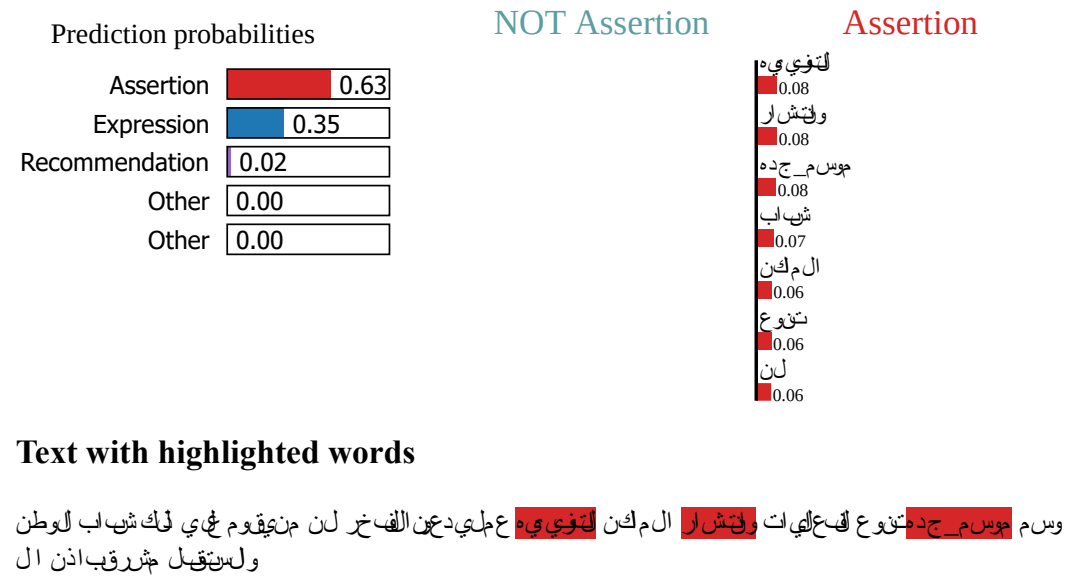

**(d)** The actual label "Expression" and predicted as "Assertion"

*Figure 5 Examples of ASAD classification errors*

## *6.2. ArSAS*

We analyzed the misclassification caused by the best model "camelbert-mix" by generating the confusion matrix in Table 7. We observed that all classes made errors with the expression class, which, again, due to the high imbalance issue. The greatest number of misclassification were found in tweets classified as assertions but predicted as expressions, as well as tweets classified as expressions and predicted as assertions. These misclassification could be the cause of the dataset's nature. Table 8 displays some misclassified examples that are classified as expressions. Figures 6a to 6d show the LIME analysis results.

*Table 7 Confusion matrix for the best model "camelbert-mix" on the ArSAS*

| | **Ass** | **Exp** | **Mis** | **Que** | **Req** |
|---|---|---|---|---|---|
| **Ass** | *1444* | **158** | 6 | 14 | 2 |
| **Exp** | **151** | *2107* | 17 | 30 | 9 |
| **Mis** | 11 | **13** | *7* | 0 | 1 |
| **Que** | 8 | **66** | 0 | *75* | 1 |
| **Req** | 2 | **22** | 0 | 1 | *10* |

*Table 8 . Examples of ArSAS classification errors*

| Tweet | Orig SA | pred SA |
|---|---|---|
| جنوب افريقيا والسنغال ضمن تصفيات كاس العالم مع التقدير والاحترام | Ass | Exp |
| يعيال انتهت تصفيات كاس العالم ولا باقي | Qus | Exp |
| اتمني محمد صلاح يروح برشلونه عشان يغطي علي ميسي رابط | Req | Exp |
| ثلاث منتخبات لازم يلعبون في كاس العال بدون تصفيات البرازيل المانيا ايطاليا و تصبحون علي خير | Mis | Exp |

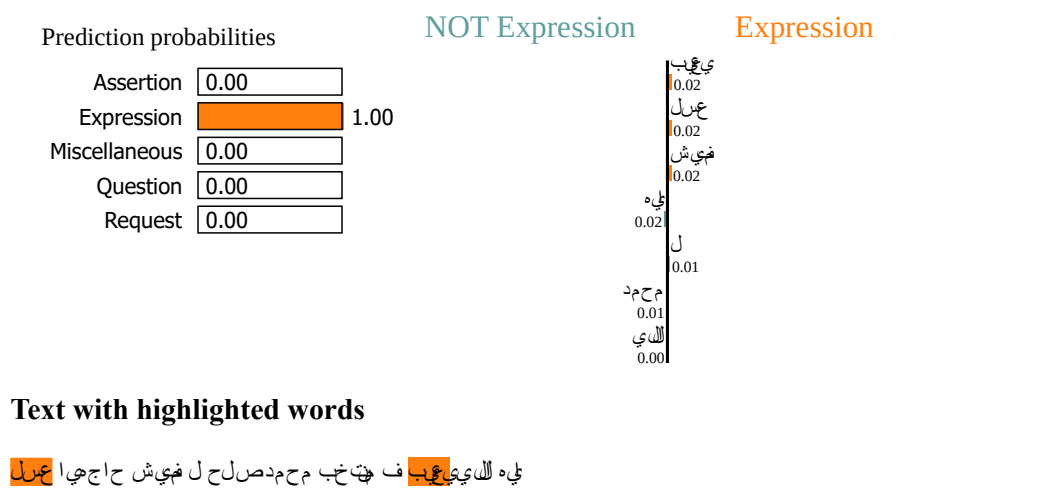

**(a)** The actual label "Expression" and the predicted label is "Expression".

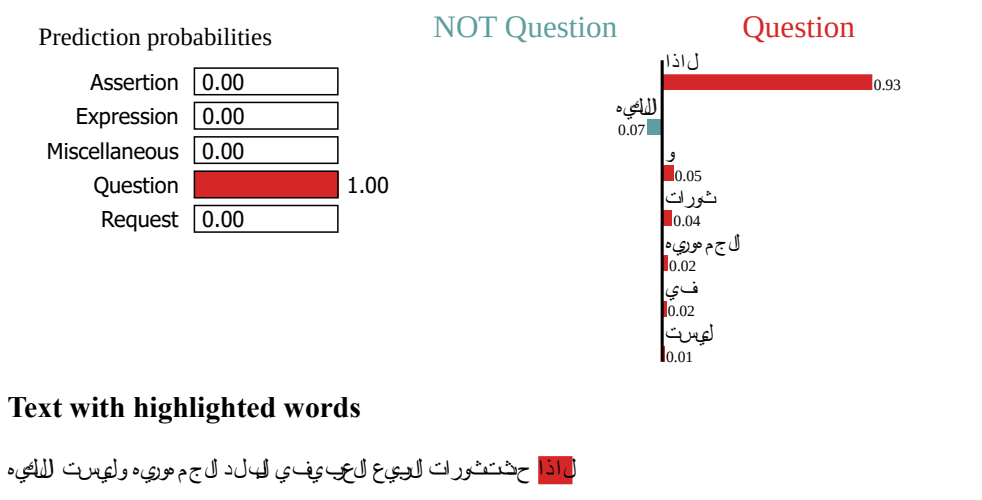

**(b)** The actual label "Question" and predicted as "Question".

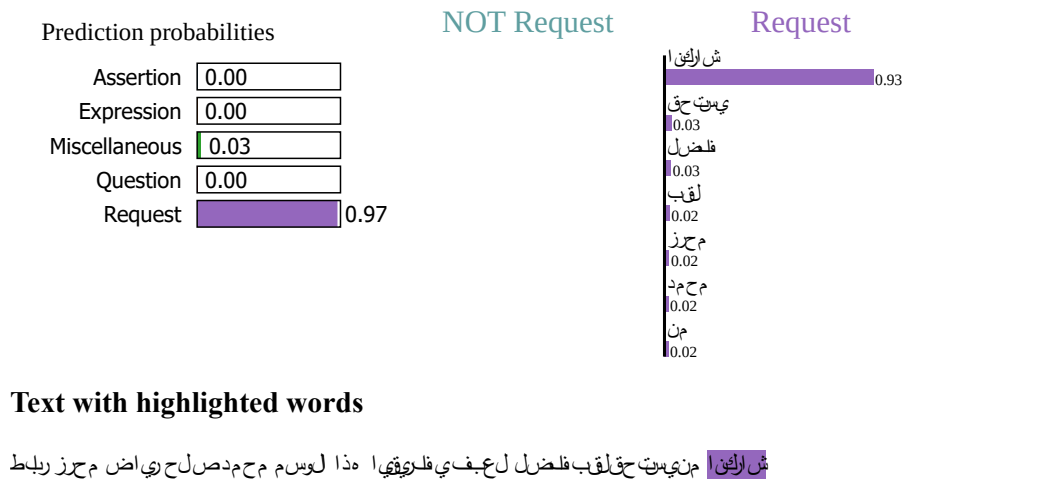

**(c)** The actual label "Question" and predicted as "Request".

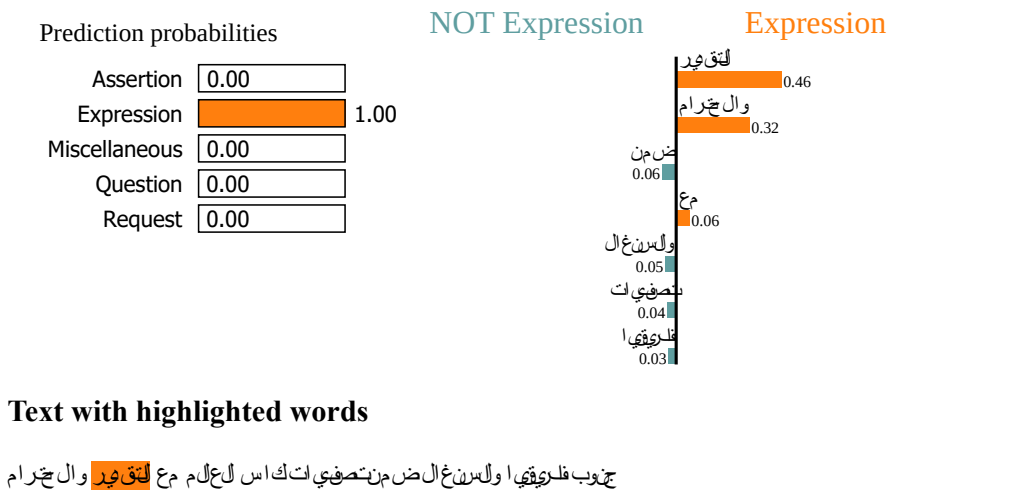

**(d)** The actual label "Assertion" and predicted as "Expression".

*Figure 6 Examples of ArSAS classification errors*

## 7. Conclusions

Speech act classification is an important task of determining the communicator's aim of a speech. Several studies have proposed methods to identify speech acts in English and other languages, and only a small number of studies have looked into Arabic Speech act classification. Thus, in this paper, we developed dialect Arabic speech act datasets that com- prise 22, 352 tweets of unspecified topics annotated with six speech act labels. We proposed a BERT-based weighted ensemble learning approach for arabic tweet act classification by combining the prediction of variant of BERT models. A comparative comparison were also performed to compare the proposed model against variant transformer-based deep neural network models and BiLSTM model in the classification of Arabic speech acts.

Our results show that the transformer-based models generally perform better than BiLSTM on the ASAD and ArSAS datasets. In addition, for all the transformer-based models, araBERTv2-Twitter and camelbert-mix perform better than the other models, with a macro-averaged F1 of 0.73 and 0.58 on the ASAD and ArSAS datasets, respectively. The results of the proposed BERT-based ensemble model have improved the overall result of ASAD with accuracy of 0.84 and 0.74 macro-averaged F1. To overcome the problem of data imbalance a transformer based data augmentation technique were implemented. The data augmentation have improved the F1 score to 0.80 and 69 for ASAD and ArSAS, respectively.

**Funding:** This research received no external funding

**DataAvailabilityStatement:**
The ASAD dataset is avalaible in the GitHub repository https://github.com/alshehrikhadejaa/ASAD.git where as the ArSAS dataset is avalabile in https://homepages.inf.ed.ac.uk/wmagdy/resources.htm.

**Conflicts of Interest:** The authors declare no conflict of interest.

## Abbreviations

The following abbreviations are used in this manuscript:

| | |
|---|---|
| MSA | Modern Standard Arabic |
| NLP | Natural Language Processing |
| ASAD | Arabic Sentiment Analysis Dataset |
| RNN | Recurrent Neural Network |
| LSTM | Long Short-Term Memory |
| BiLSTM | Bidirectional Long Short-Term Memory |
| BERT | Bidirectional Encoder Representations from Transformers |
| RNN | Recurrent Neural Network |
| SVM | Support vector machine |
| CNN | Convolutional Neural Network |
| ArSAS | Arabic Speech-Act and Sentiment Corpus |

## Author Contributions

Conceptualization, K.A., A.A. and N.A.; methodology, K.A., A.A. and A.A.; software, K.A.; validation, K.A., A.A. and N.A.; investigation K.A., A.A. and N.A.; writing—original draft, K.A.; writing—review and editing A.A. and N.A.; supervision, A.A. and N.A.; funding acquisition, none. All authors have read and approved the submission of this paper. All authors have read and agreed to the published version of the manuscript.